\documentclass[a4paper, 10pt, conference]{IEEEtran}      
\usepackage{FG2017}

\FGfinalcopy 

\usepackage{verbatim}
\IEEEoverridecommandlockouts                              

\usepackage{graphics} 
\usepackage{epsfig} 
\usepackage{mathptmx} 
\usepackage{times} 
\usepackage{amsmath} 
\usepackage{amssymb}  

\title{\LARGE \bf
Towards Estimating the Upper Bound of Visual-Speech Recognition:
\hspace{15mm} The Visual Lip-Reading Feasibility Database
}

\author{\parbox{16cm}{\centering
    {\large Adriana Fernandez-Lopez, Oriol Martinez and Federico M. Sukno}\\
    {\normalsize
    Department of Information and Communication Technologies, Pompeu Fabra University, Barcelona, Spain\\}}
}

\begin{document}
\IEEEoverridecommandlockouts\IEEEpubid{\makebox[\columnwidth]{978-1-5090-4023-0/17/\$31.00~\copyright{}2017 IEEE \hfill}
\hspace{\columnsep}\makebox[\columnwidth]{ }}
\ifFGfinal
\thispagestyle{empty}
\pagestyle{empty}
\else
\author{Anonymous FG 2017 submission\\-- DO NOT DISTRIBUTE --\\}
\pagestyle{plain}
\fi
\maketitle

\bstctlcite{IEEEexample:BSTcontrol}

\begin{abstract}
Speech is the most used communication method between humans and it involves the perception of auditory and visual channels. Automatic speech recognition focuses on interpreting the audio signals, although the video can provide information that is complementary to the audio. Exploiting the visual information, however, has proven challenging. On one hand, researchers have reported that the mapping between phonemes and visemes (visual units) is one-to-many because there are phonemes which are visually similar and indistinguishable between them. On the other hand, it is known that some people are very good lip-readers (e.g: deaf people).
We study the limit of visual only speech recognition in controlled conditions. With this goal, we designed a new database in which the speakers are aware of being read and aim to facilitate lip-reading. In the literature, there are discrepancies on whether hearing-impaired people are better lip-readers than normal-hearing people. Then, we analyze if there are differences between the lip-reading abilities of 9 hearing-impaired and 15 normal-hearing people. Finally, human abilities are compared with the performance of a visual automatic speech recognition system. In our tests, hearing-impaired participants outperformed the normal-hearing participants but without reaching statistical significance. Human observers were able to decode 44\% of the spoken message. In contrast, the visual only automatic system achieved 20\% of word recognition rate. However, if we repeat the comparison in terms of phonemes both obtained very similar recognition rates, just above 50\%. This suggests that the gap between human lip-reading and automatic speech-reading might be more related to the use of context than to the ability to interpret mouth appearance.
\end{abstract}

\section{INTRODUCTION}
\label{sec:intro}
Speech is the most used communication method between humans, and it is considered a multi-sensory process that involves perception of both acoustic and visual cues since McGurk demonstrated the influence of vision in speech perception. Many authors have subsequently demonstrated that the incorporation of visual information into speech recognition systems improves their robustness \cite{mcgurk1976hearing, potamianos2003recent}.

Visual information usually involves position and movement of the visible articulators (the lips, the teeth and the tongue), speaker localization, articulation place and other signals not directly related to the speech (facial expression, head pose and body gestures) \cite{hilder2009comparison, williams1998frame, chictu2012automatic}. Even though the audio is in general much more informative than the video signal, speech perception relies on the visual information to help decoding spoken words as auditory conditions are degraded \cite{hilder2009comparison, erber1975auditory, sumby1954visual, ronquest2010language}. Furthermore, for people with hearing impairments, the visual channel is the only source of information to understand spoken words if there is no sign language interpreter \cite{potamianos2003recent, antonakos2015survey, seymour2008comparison}. Therefore, visual speech recognition is implicated in our speech perception process and is not only influenced by lip position and movement but it also depends on the speaker's face, as it has been shown that it can also transmit relevant information about the spoken message \cite{williams1998frame, chictu2012automatic}.
Much of the research in Automatic Speech Recognition (ASR) systems has focused on audio speech recognition, or on the combination of both modalities using Audio-Visual Automatic Speech Recognition (AV-ASR) systems to improve the recognition rates, but Visual Automatic Speech Recognition (VASR) systems have been less frequently analyzed alone \cite{dupont2000audio, nefian2002coupled, zhou2014review, yau2007visual, sui2015listening, chung2016lip,petridis2016deep, almajai2016improved}. The performance of audio only ASR systems is very high if there is not much noise to degrade the signal. However, in noisy environments AV-ASR systems improves the recognition performance when compared to their audio-only equivalents \cite{potamianos2003recent, dupont2000audio}. In contrast, in visual only ASR systems the recognition rates are rather low \cite{zhou2014compact}. This can be partially explained by the higher difficulty associated to decoding speech through the visual channel, when compared to the audio channel.

One of the key limitations of VASR systems resides on the ambiguities that arise when trying to map visual information into the basic phonetic unit (phonemes), i.e. not all the phonemes that are heard can be distinguished by observing the lips. There are two types of ambiguities: $i)$ there are phonemes that are easily confused because they look visually similar between them (e.g: /p/, /b/ and /m/). For example, the phones /p/ and /b/ are visually indistinguishable because voicing occurs at the glottis, which is not visible; $ii)$ there are phonemes whose visual appearance can change (or even disappear) depending on the context. This is the case of the \emph{velars}, consonants articulated with the back part of the tongue against the soft palate (e.g: /k/ or /g/), because they change their position in the palate depending on the previous or following phoneme. Specifically, velar consonants tolerate palatalization (the phoneme changes to palatal) when the previous or following phoneme is a vowel or a palatal \cite{moll1971investigation}. Other drawbacks associated to lipreading have also been reported in the literature, such as the distance between the speakers, illumination conditions or visibility of the mouth \cite{hilder2009comparison, buchan2007spatial, ortiz2008lipreading}. However, the latter can be easily controlled, while the ambiguities explained above are limitations intrinsic to lip-reading and constitute an open problem.

On the other hand, it is known that some people are very good lip-readers. In general, visual information is the only source of reception and comprehension of oral speech for people with hearing impairments, which leads to the common misconception that they must be good lip-readers. Indeed, while many authors have found evidence that people with hearing impairments outperform normal-hearing people in comprehending visual speech \cite{potamianos2001automatic, bernstein1998makes, capek2008cortical, ellis2001tas, lyxell2000visual}, there are also several studies where no differences were found in speech-reading performance between normal-hearing and hearing-impaired people \cite{rodriguez2015speechreading, kyle2013speechreading}. Such conflicting conclusions might be partially explained by the influence of other factors beyond hearing impairment. For example, it is well know that  human lip-readers use the context of the conversation to decode the spoken information \cite{hilder2009comparison, chictu2012automatic, buchan2007spatial}, thus it has been argued that people who are good lip-readers might be more intelligent, with more knowledge of the language, and with a more comprehensible oral speech for others \cite{ortiz2008lipreading, rodriguez2015speechreading, mohammed2006speechreading, kyle2006concurrent}.

While the above complexities may provide some explanation to the rather low recognition rates of VASR systems, there seems to be a significant gap between these and human lip-reading abilities. More importantly, it is not clear what would be the upper bound of visual-speech recognition, especially for systems not using context information (it has been argued that humans can \emph{read} only around $30\%$ of the information from the lips, and the rest is filled-in from the context \cite{ortiz2008lipreading, duchnowski2000development}). Thus, it is not clear if the poor recognition rates of VASR systems are due to inappropriate or incomplete design or because there is an intrinsic limitation in visual information that causes the impossibility of perfect decoding of the spoken message.

\textbf{Contributions: }In this work we explore the feasibility of visual speech reading with the aim to estimate the recognition rates achievable by human observers under favorable conditions and compare them with those achieved by an automatic system. To this end, we focus on the design and acquisition of an appropriate database in which recorded speakers actively aim to facilitate lip-reading but conversation context is minimized. Specifically, we present a new database recorded with the explicit goal of being visually informative of the spoken message. Thus, data acquisition is especially designed with the aim that a human observer (or a system) can decode the message without the help of the audio signal. Concretely, lip-reading is applied to people that is aware of being read and has been instructed to make every effort so that they can be understood based exclusively on visual information. Then, the database deals with sentences that are uttered slowly, with repetitions, well pronounced and viewed under optimal conditions ensuring good illumination and mouth visibility (without occlusions and distractions).

In this database we divided the participants in two groups: 9 hearing-impaired subjects and 15 normal-hearing subjects. In our tests, hearing-impaired participants outperformed the normal-hearing participants but without reaching statistical significance. Human observers outperform markedly the VASR system in terms of word recognition rates, but in terms of phonemes, the automatic system achieves very similar accuracy to human observers.

\section{Audio-visual speech databases}
\label{sec:databases}
Visual only speech recognition spans over more than thirty years, but even today is still an open problem in science. One of the limitations for the analysis of VASR systems is the accessible data corpora. Despite the abundance of audio speech databases, there exist a limited number of databases for audio-visual or visual only ASR research. That is explained in the literature because the field is relatively young, and also, because the audio-visual databases add some challenges such as database collection, storage and distribution, not found as a problem in audio corpora. Acquisition of visual data at high resolution, frame rate and image quality, with optimal conditions and synchronized with the audio signal requires expensive equipment. In addition, visual storage is at least one or two orders of magnitude to the audio signal, making his distribution more difficult \cite{zhou2014review}, \cite{potamianos2004audio}.

Most databases used in audio-visual ASR systems suffer from one or more weaknesses. For example, they contain low number of subjects (\cite{matthews2002extraction, cox2008challenge}), small duration (\cite{matthews2002extraction, cox2008challenge, lee2004avicar, messer1999xm2vtsdb}), and are addressed to specific and simple recognition tasks. For instance, most corpora are centered in simple tasks such as isolated or connected letters (\cite{matthews2002extraction, cox2008challenge, lee2004avicar}), digits (\cite{lee2004avicar, messer1999xm2vtsdb, patterson2002cuave, huang2004audio, lucey2008patch}), short sentences (\cite{messer1999xm2vtsdb, sanderson2002vidtimit, cooke2006audio, mccool2012bi, zhao2009lipreading, anina2015ouluvs2}) and only recently continuous speech (\cite{huang2004audio, ortega2004av, hazen2004segment, LiILiR}). These restrictions make more difficult the generalization of methods and the construction of robust models because of the few samples of training. Additional difficulties are that some databases are not freely available.

As explained in Section \ref{sec:intro} the aim of this project is to apply continuous lip-reading to people that is conscious of being read and is trying to be understood based exclusively on visual information. Thus, from the most common databases, only VIDTIMIT \cite{sanderson2002vidtimit}, AVICAR \cite{lee2004avicar}, Grid \cite{cooke2006audio}, MOBIO \cite{mccool2012bi}, OuluVS \cite{zhao2009lipreading}, OuluVS2 \cite{anina2015ouluvs2}, AV@CAR \cite{ortega2004av}, AV-TIMIT \cite{hazen2004segment}, LILiR \cite{LiILiR} contain short sentences or continuous speech and could be useful to us. However, we rejected the use of them because the participants speak in normal conditions without previous knowledge of being lip-read. In addition, most of the databases have low technical aspects and limited number of subjects with restricted vocabularies centred in repetitions of short utterances. Subsequently, we decided to develop a new database designed specifically for recognizing continuous speech in controlled conditions.

\section{Visual Lip-Reading Feasibility Database}
\label{sec:VLRF}

The Visual Lip-Reading Feasibility (VLRF) database is designed with the aim to contribute to research in visual only speech recognition. A key difference of the VLRF database with respect to existing corpora is that it has been designed from a novel point of view: instead of trying to lip-read from people who are speaking naturally (normal speed, normal intonation,...), we propose to lip-read from people who strive to be understood.

Therefore, the design objective was to create a public database visually informative of the spoken message in which it is possible to directly compare human and automatic lip-reading performance. For this purpose, in each recording session there were two participants: one speaker and one lip-reader. The speaker was recorded by a camera while pronouncing a series of sentences that were provided to him/her; the lip-reader was located in a separate room, acoustically isolated from the room where the speaker was located. To make the human decoding as close as possible to the automatic decoding, the input to the lip-reader was exclusively the video stream recorded by the camera, which was displayed in real time by means of a 23" TV screen.

After each uttered sentence, the lip-reader gave feedback to the speaker (this was possible because it was possible to enable audio feedback from the lip-reading room to the recording room, but not conversely). Each sentence could be repeated up to 3 times, unless the lip-reader decoded it correctly in fewer repetitions. Both the speaker utterances and the lip-reader answers (at each repetition) were annotated.

Participants were informed about the objective of the project and the database. They were also instructed to make their best effort to be easily understood, but using their own criteria (e.g: speak naturally or slowly, emphasize separation between words, exaggerate vocalization,...).

Each recording session was divided in 4 levels of increasing difficulty: 3 levels with 6 sentences and 1 level with 7 sentences. We decided to divide the session in different levels to make it easier for participants to get accustomed to the lip-reading task (and perhaps also to the speaker). Specifically, in the first level the sentences are short with only few words, and as the level increases the difficulty increases in terms of number of words. The sentences are unrelated among them and only the context within the sentence is present. Thus, in the first sentences participants had to read fewer words but with very little context and in the last sentences the context was considerably more important and would certainly help decoding the sentence. To motivate participants and to ensure their concentration during all the session, at the end of each level both participants changed their roles.

Finally, because our objective was to determine the visual speech recognition rates that could be achievable, we also recruited volunteers which were hearing-impaired and accustomed to use lip-reading in their daily routine. Then, we will also compare the capability of lip-reading of normal-hearing and hearing-impaired people.

\subsection{Participants}
We recruited 24 adult volunteers (3 male and 21 female). Thirteen are University students, one is Teacher of Sign Language at UPF and the other 10 participants are members of the Catalan Federation of Associations of Parents and Deaf (ACCAPS) \cite{webACCAPS}. The 24 participants were divided in two groups: normal-hearing people and hearing-impaired people.

-- \textit{Normal-hearing participants.}
Fifteen of the volunteers are normal-hearing participants (14 females and 1 male), who were selected from a similar educational range (e.g: same degree) because, as explained in Section \ref{sec:intro}, lip-reading abilities have been related to intelligence and language knowledge. Two of the participants were more than 50 years old and have a different education level while the other 13 subjects of this group shared educational level and age range.

-- \textit{Hearing-impaired participants.}
There were nine hearing-impaired participants, all above 30 years old (7 female and 2 male). Eight of them have post-lingual deafness (the person loses hearing after acquiring spoken language) and one has pre-lingual deafness (the person loses hearing before the acquisition of spoken language). There were 4 participants with cochlear implants or hearing aids.

\subsection{Utterances}
Each participant was asked to read 25 different sentences, from a total pool of 500 sentences, proceeding similarly to \cite{cooke2006audio}. The sentences were unrelated between them to avoid that lip-readers could benefit from conversation context. Sentences had different levels of difficulty, in terms of their number of words. There were 4 different levels, from 3-4 words, 5-6 words, 7-8 words and 8-12 words. We decided to divide the sentences in different levels for two reasons. Firstly, to allow lip-readers to get some \emph{training} with the short sentences of the first level (i.e. to get acquainted and gain confidence with the setup, the task and the speaker). Secondly, to compare the effect of the context in the performance of human lip-readers. The utterances with fewer words have very little context, while longer sentences contained considerable context that should help the lip-reader when decoding the message.

Overall, there were 10200 words in total (1374 unique), with an average duration of 7 seconds per sentence and a total database duration of 180 minutes (540,162 frames). The sentences contained a balanced phonological distribution of the Spanish language, based on the balanced utterances used in the AV@CAR database \cite{ortega2004av}.

\subsection{Technical aspects}

\begin{figure}[t]
        \centering
        \includegraphics[width=\linewidth, height=4cm]{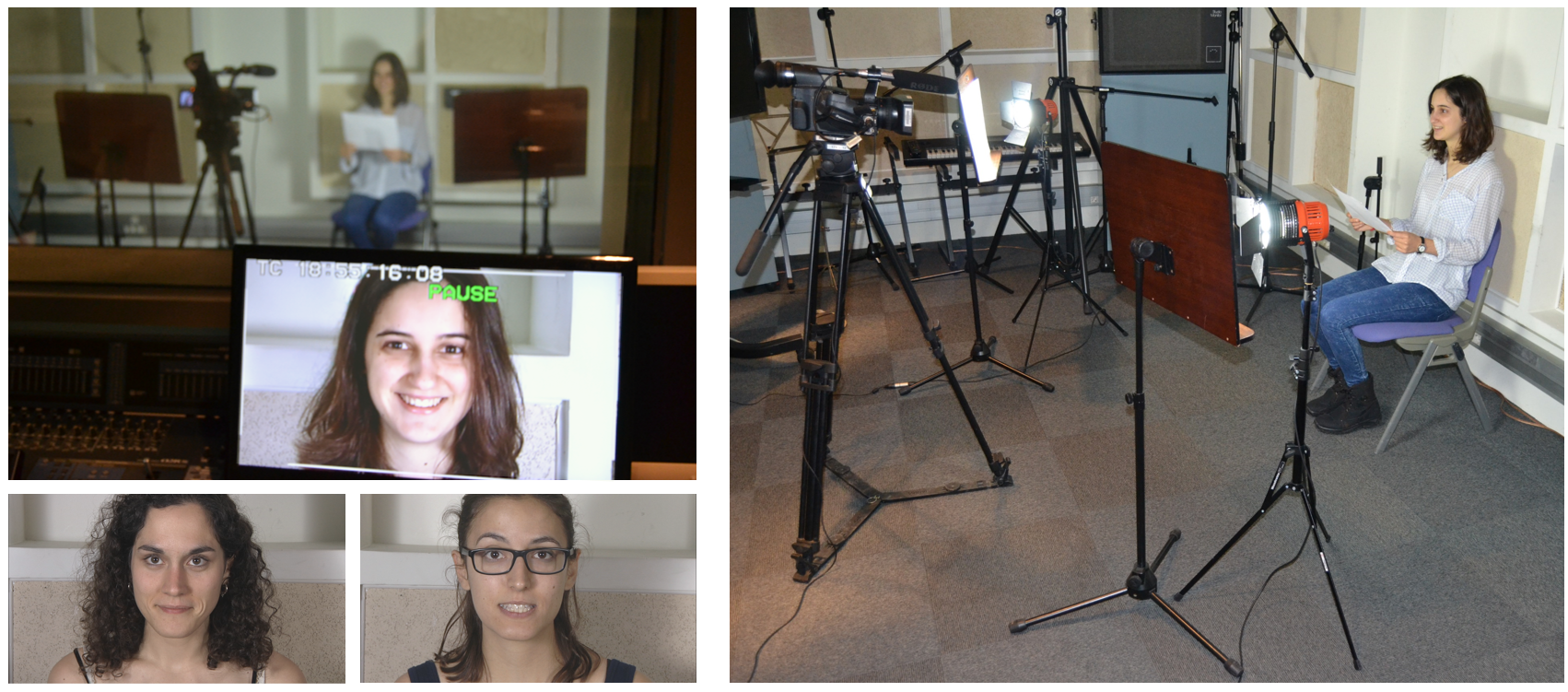}
        \caption{Scheme of the recording setup and snapshots of the VLRF database.}
        \label{fig:roomDistribution}
\end{figure}

The database was recorded in two contiguous soundproof rooms (Fig. \ref{fig:roomDistribution}). The distribution of the recording equipment into the rooms is shown in Fig. \ref{fig:roomDistribution}. A Panasonic HPX 171 camera was located with a tripod PRO6-HDV in front of the chair of the speaker, to ensure an approximately frontal face shot, with a supplementary directional microphone mounted on the camera to ensure a directional coverage in the direction of the speaker. The camera recorded a close up shot (Fig.\ref{fig:roomDistribution}) at 50 fps with a resolution of $1280 \times 720$ pixels and audio at 48 kHz mono with 16-bit resolution. Two Lumatek ultralight 1000W Model 53-11 were used together with reflecting panels to obtain a uniform illumination and minimize shadows or other artifacts on the speaker's face.
When performing the lip-reading task, the lip-reader was located in the control room. The position of the lip-reader was just in front of a 23" LG Flatron M2362D PZ TV. This screen was connected to the camera so that it reproduced in real time what the camera was recording. Only the visual channel of the camera was fed into the control room, although both audio and video channels are recorded for post processing of the database.
The rooms were acoustically isolated between them except for the feedback channel composed by a microphone in the control room and a loudspeaker in the recording room. This channel was used after each utterance to let the speaker know what message was decoded by the lip-reader.

\subsection{Data labeling}

The ground-truth of the VLRF database consists of a phoneme label per frame. We used the EasyAlign plug-in from Praat \cite{boersma2002praat}, which allows to locate the phoneme in each time instant based on the audio stream. Specifically, the program locates the phonemes semi-automatically and there is usually the need for manual intervention to adapt the boundaries of each phoneme to more precise positions. The phonemes used are based on the phonetic alphabet SAMPA \cite{wells1997sampa}. For the Spanish language, the SAMPA vocabulary is composed of the following 31 phonemes: /p/, /b/, /t/, /d/, /k/, /g/, /tS/, /jj/, /f/, /B/, /T/, /D/, /s/, /z/, /x/, /G/, /m/, /n/, /N/, /J/, /l/, /L/, /r/, /4/, /j/, /w/, /a/, /e/, /i/, /o/, /u/.

\section{Results}
\label{sec:Results}
In this section we show the word- and phoneme-recognition rates obtained in our experiments. We start by analyzing the human lip-reading abilities and comparing the performance of hearing-impaired and normal-hearing participants. Then, we analyse the influence of training and context in human performance. Finally, we compare the performance of our automatic system to the results obtained by human observers.

The use of two separate measures (word and phoneme rates) is necessary to analyze different aspects of our results. On one hand, phonemes are the minimum distinguishable units of speech and directly constitute the output of our automatic system. However, the ultimate goal of lip-reading is to \emph{understand} the spoken language, hence the need to focus (at least) on words. It is important to notice that acceptable phoneme recognition rates do not necessarily imply good word recognition rates, as will be shown later.

The word recognition rate was computed as the fraction of words correctly understood in a given sentence. The phoneme recognition rate was computed as the fraction of video frames in which the correct phoneme was assigned. Consequently, 25 accuracy measures were computed for each participant and each repetition. Recognition rates for the automatic system were computed in the same manner, except that there were no multiple repetitions.

\subsection{Experimental setup}

Our VASR system starts by detecting the face and performing an automatic location of the facial geometry (landmark location) using the Supervised Descend Method (SDM) \cite{xiong2013supervised}. Once the face is located, the estimated landmarks are used to fix a bounding box around the  region (ROI) that is then normalized to a fixed size. Later on, local appearance features are extracted from the ROI based on early fusion of DCT and SIFT descriptors in both spatial and temporal domains. As explained in Section \ref{sec:intro} there are phonemes that share the same visual appearance and should belong to the same class (visemes). Thus, we constructed a phoneme to viseme mapping that groups 32 phonemes into 20 visemes based on an iterative process that computes the confusion matrix and merges at each step the phonemes that show the highest ambiguity until the desired length is achieved. Then, the classification of the extracted features into phonemes is done in two steps. Firstly, multiple LDA classifiers are trained to convert the extracted features into visemes and secondly, at the final step, one-state-per-class HMMs are used to model the dynamic relations of the estimated visemes and produce the final phoneme sequences. This system was shown to produce near state-of-the-art performance for continuous visual speech-reading tasks (more details in \cite{fernandez2016Automatic}).

\subsection{Human lip-reading}

\begin{figure}[tb]
    \centering
    \includegraphics[width=9cm, height=2.7cm]{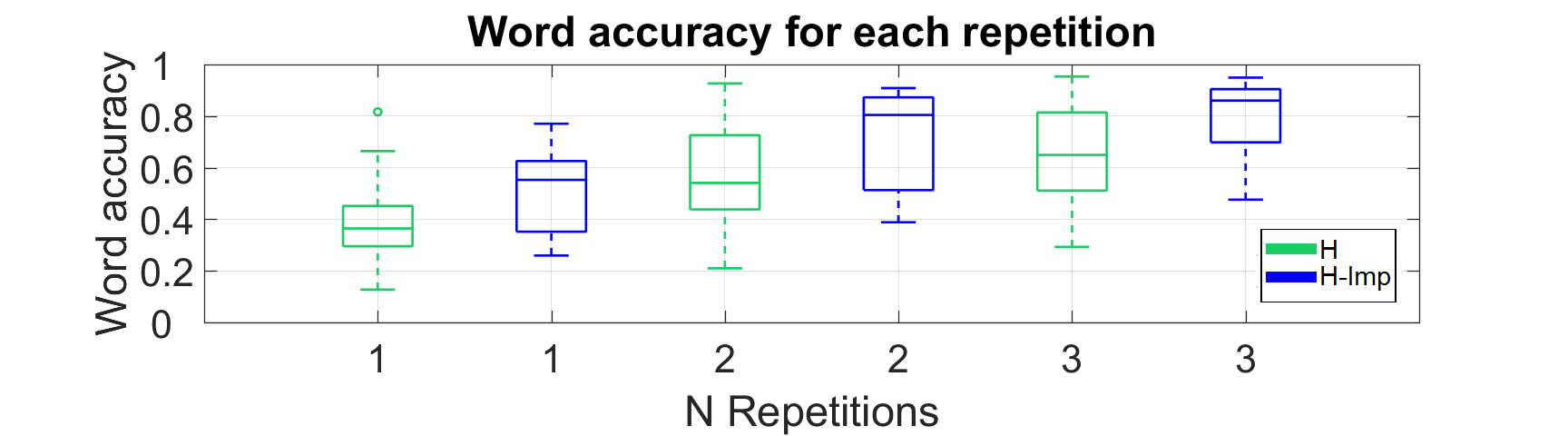}
    \includegraphics[width=9cm, height=2.7cm]{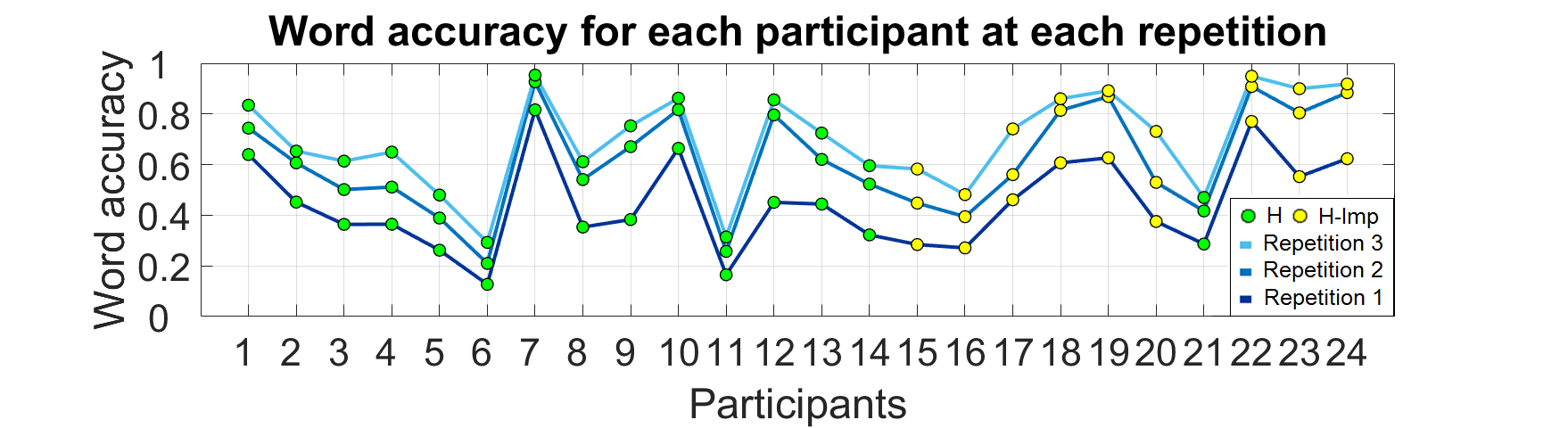}
    \caption{Top: word accuracy for normal-hearing (H) and hearing-impaired groups (H-Imp) at each repetition; Bottom: word accuracy per participant at each repetition.}
    \label{fig:WaccUsers}
\end{figure}

As explained in Section \ref{sec:intro}, it is not clear if hearing-impaired people are better lip-readers than normal-hearing people.
Fig. \ref{fig:WaccUsers} (\textit{Top}) shows the word recognition rates for both groups at each repetition and Fig. \ref{fig:WaccUsers} (\textit{Bottom}) shows the word recognition rates for each participant and repetition. Analyzing each participant individually, it is difficult to observe any group-differences between hearing-impaired and normal-hearing participants. However, we do observe large performance variations within each of the groups, i.e. there are very good and quite poor lip-readers regardless of their hearing condition.

On the other hand, looking at the results globally, split only by group (Fig. \ref{fig:WaccUsers} (\textit{Top})), they suggest that hearing-impaired participants outperform normal-hearing participants in the lip-reading task for all three repetitions. However, the results differ about 20\% in terms of word recognition rate and thus we need to study if this difference is statistically significant.

To do so, we estimated the word accuracy of each participant as the average accuracy across the 25 sentences that he/she had to lip-read. Then, we performed statistical tests to determine if there were significant differences between the 9 hearing-impaired samples and the 15 normal-hearing samples.
Because we only want to test if the hearing-impaired participants were better than normal-hearing participants, we performed single-tailed tests where the null hypothesis was that the mean or median (depending on the test) performance of hearing-impaired participants was not higher than the performance of normal-hearing participants. We ran two tests (summarized in Table \ref{tab:StatisticalTests}) for each of the 3 repetitions: \textit{Wilcoxon signed rank test} and \textit{Unpaired two-sample t-test}. Taking the conventional significance threshold of $p < 0.05$ it could be argued that at the third repetition the performance of hearing-impaired participants was significantly better than that of normal-hearing participants. However, this was not observed in the first two repetitions. Moreover, the 9 hearing-impaired subjects did better than the 15 normal-hearing, but taking into account that the sample size is relatively small, current trends in statistical analysis suggest that the obtained p-values are not small enough to claim that this would extrapolate to the general population. On the other hand, looking at the p-values, with the current number of subjects we are not far from reaching significance \cite{colquhoun2014investigation}.

\begin{table}[tb]
\begin{center}
\caption{Statistical comparison between hearing-impaired and normal-hearing participants at each repetition.}
\label{tab:StatisticalTests}
\begin{tabular}{|c|p{2.8cm}|p{2.8cm}|}
 \hline
  \textbf{Attempt} & \textbf{Wilcoxon signed rank} &  \textbf{Unpaired two-sample} \\
   \hline\hline
   1 & p = 0.116 & p = 0.094 \\
 \hline
   2 & p = 0.094 & p = 0.088 \\
  \hline
  3 & p = 0.041 &  p = 0.037 \\
 \hline
\end{tabular}
\end{center}
\end{table}

In Fig. \ref{fig:WaccUsers} we also show the influence of repetitions into the final performance: as the number of repetitions increases the recognition rate increases too. This effect can be seen split by group and analysing each participant separately.

\subsection{Training and context influence on lip-reading}

The context is one of the human resources more used in lip-reading to complete the spoken message. To analyse the influence of the context, the participants were asked to read four different types of sentences, in terms of number of words (explained in Section \ref{sec:VLRF}). Thus, as the level increases, sentences are longer and the context increases too.

\begin{figure}
      \centering
        \includegraphics[width=8cm, height=2.7cm]{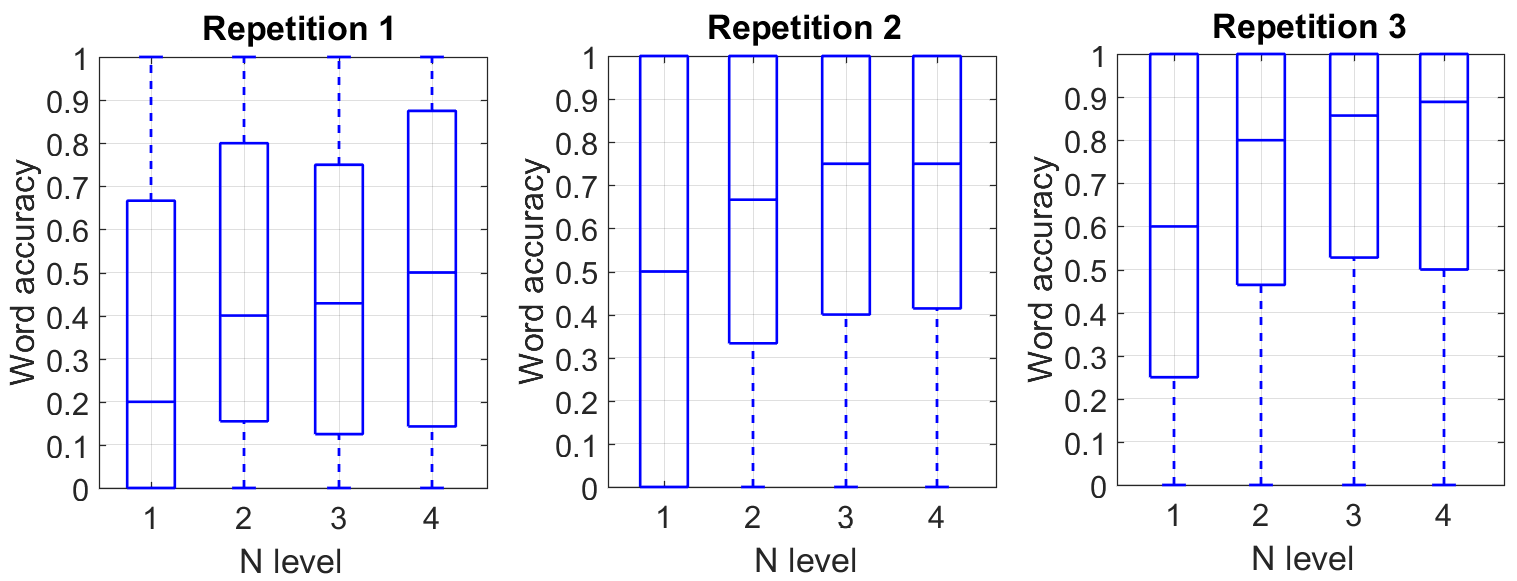}
        \caption{Word recognition average for each participant at each level.}
        \label{fig:Wacclevels}
\end{figure}

In Fig. \ref{fig:Wacclevels} we can observe how the first level has the lowest word recognition rates for all repetitions, while the last level has the highest rates. There are two factors that could contribute to this effect: 1) Context: humans use the relation between words to try decoding a meaningful message, and 2) Training: as the level increases the participants are more acquainted to the speaker and to the lip-reading task.

\begin{figure}[tb]
    \centering
    \includegraphics[width=9cm, height=2.7cm]{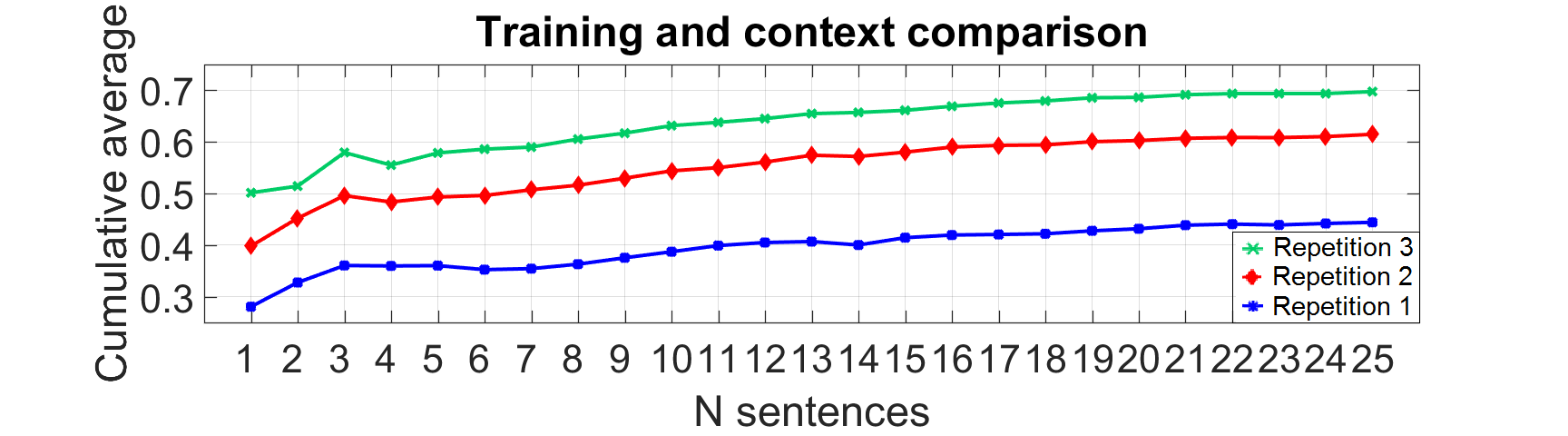}
    \caption{Cumulative average per sentence for all participants at each repetition.}
    \label{fig:HearingDeaf}
\end{figure}

The results of Fig. \ref{fig:Wacclevels} are not enough to determine whether the effect is due to context, training or both. Thus, in Fig. \ref{fig:HearingDeaf} we analyze the variation of performance per sentence (with a cumulative average) instead of per level, which should make clearer the effect of training. This is because training occurs continuously from one sentence to another while context only increases when we change from one level to the next one. Thus, the effect of training can be seen as the constant increase performance in each of the curves (up to 20\%). As the users have lip-read more sentences they tend to become better lip-readers. On the other hand, the influence of context is better observed by comparing the different repetitions. In the first attempt, the sentence was completely unknown to the participants, but, in the second and third repetitions there was usually some context available because the message had been already partially decoded, hence constraining the possible words to complete the sentence.

\subsection{Human observers and automatic system comparison}

\begin{figure}[bt]
    \centering
    \includegraphics[width=9cm, height=2.7cm]{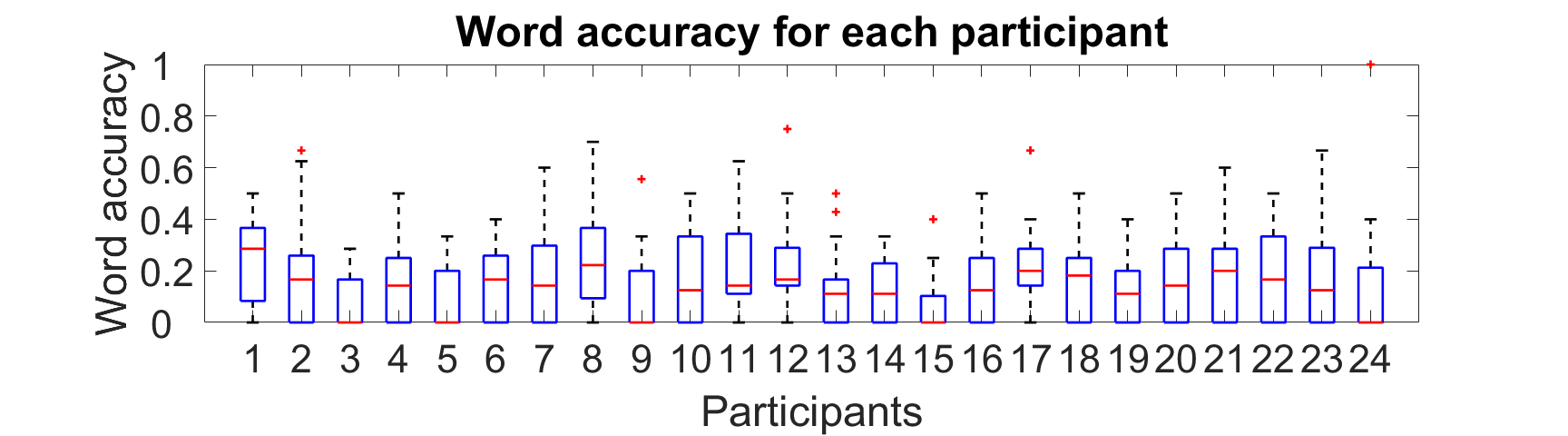}
    \includegraphics[width=9cm, height=2.7cm]{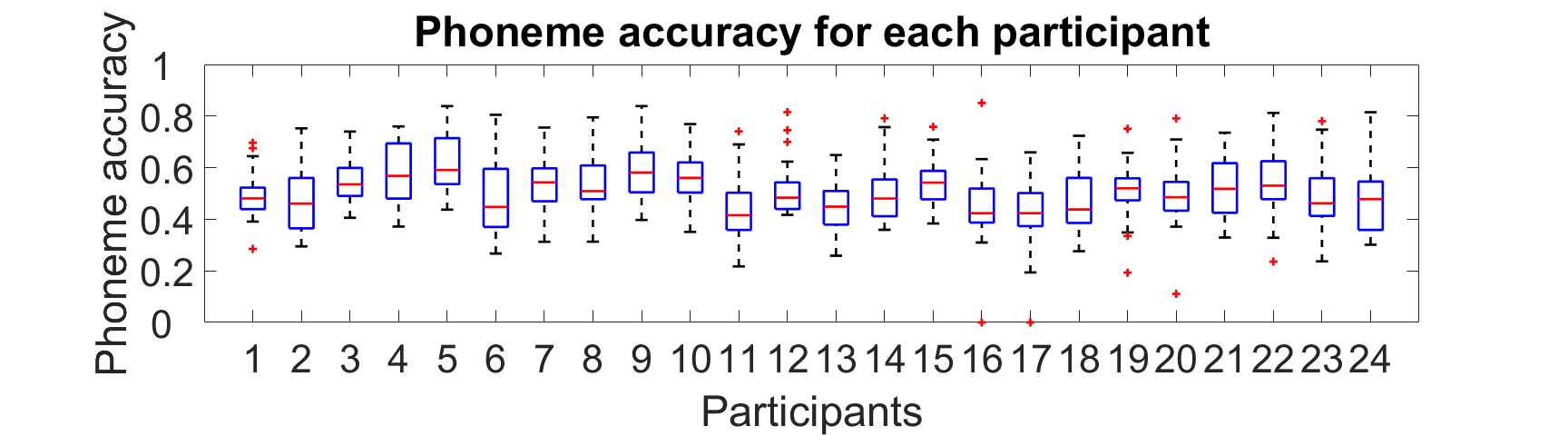}
    \caption{Top: system performance in terms of word recognition rate for each participant. Bottom: system performance in terms of phoneme recognition rate for each participant.}
    \label{fig:WaccAllUsersSystem}
\end{figure}

The results of the automatic system are only computed for the first attempt, since it was not designed to benefit from repetitions. The resulting word-recognition rates are shown in Fig. \ref{fig:WaccAllUsersSystem} (\textit{Top}). Notice that now the participant number indicates the person that was pronouncing the sentences as the recognition is always performed by the system. Thus, this figure provides information about how well the system was able to lip-read each of the participants. The system produced the highest recognition rates for participants 1, 8, 17 and 21. Interestingly, these participants had good pronunciation and visibility of the tongue and teeth.

\begin{figure}[ht]
    \centering
    \includegraphics[width=9cm, height=2.7cm]{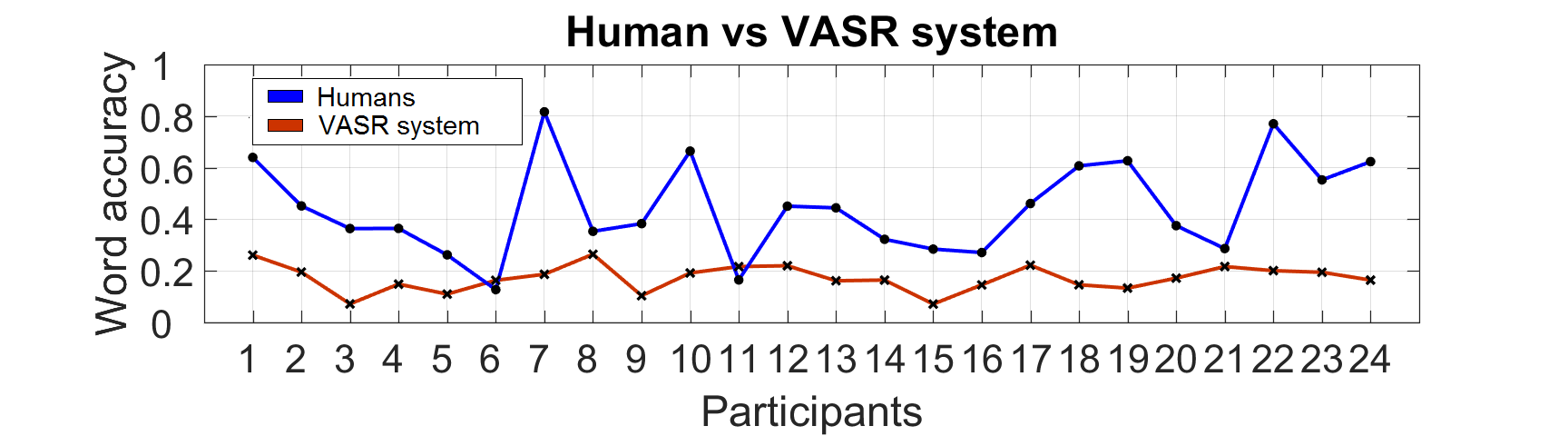}
    \includegraphics[width=9cm, height=2.7cm]{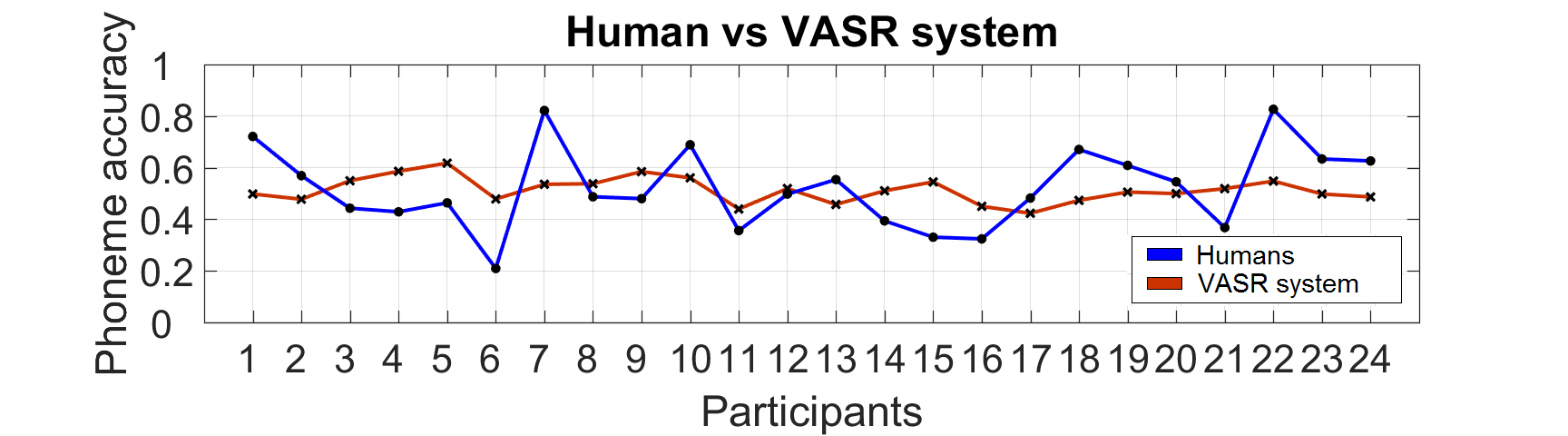}
    \caption{Top: human observers performance (Repetition 1) and automatic system performance for each participant in terms of word recognition average; Bottom: human observers performance (Repetition 1) and automatic system performance for each participant in terms of phoneme recognition average.}
    \label{fig:SystemVShuman}
\end{figure}

We are interested in comparing the performance of humans lip-reading and a VASR system. Focusing on Fig. \ref{fig:SystemVShuman} (\textit{Top}) we can observe how the word recognition rates are lower for the system in most of the cases. However, we have to take into account that the system does not use the context into the sentence. Indeed, the system is not even targeting words but phonemes, which are later merged to form words. In contrast, people directly search for correlated words with the lip movements of the speaker. Thus, it is reasonable to expect a considerable gap between human and automatic performance, which will be shown to reduce considerably if the comparison is done in terms of phonemes.

In the same figure (Fig. \ref{fig:SystemVShuman}) we can observe a direct comparison of the mean recognition rates of each participant identified by humans and by the automatic system. The system gives an unbiased measure about the facility to lip-read participants because it evaluates each of them in the same manner. In contrast, human lip-reading was performed in couples (couples are organized in successive order, e.g. participants 1 and 2, 3 and 4, etc), hence each participant was only lip-read by its corresponding partner. Analyzing Fig. \ref{fig:SystemVShuman} we can identify which users were good lip-readers and also good speakers. For example, participant 7 was lip-read by participant 8 with high word recognition rate. Then, in the curve corresponding to human performance, we observe a high value for participant 8, meaning that he/she was very successful at lip-reading. When we look at system's performance, however, the value assigned to participant 8 corresponds to the rate obtained by the system and is therefore a measure related to how participant 8 spoke rather than how he/she lip-read. For this specific participant, the figure shows that system performance was also high, hence he/she is a candidate to be good lip-reader and speaker.

\begin{figure}
      \centering
        \includegraphics[width=9cm, height=2.7cm]{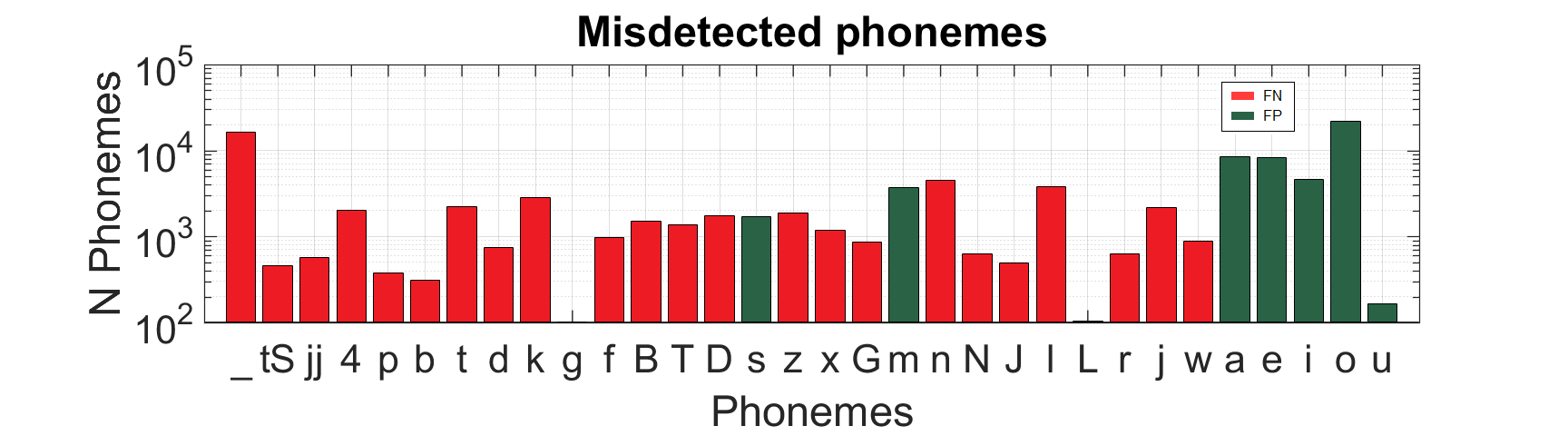}
        \includegraphics[width=9cm, height=2.7cm]{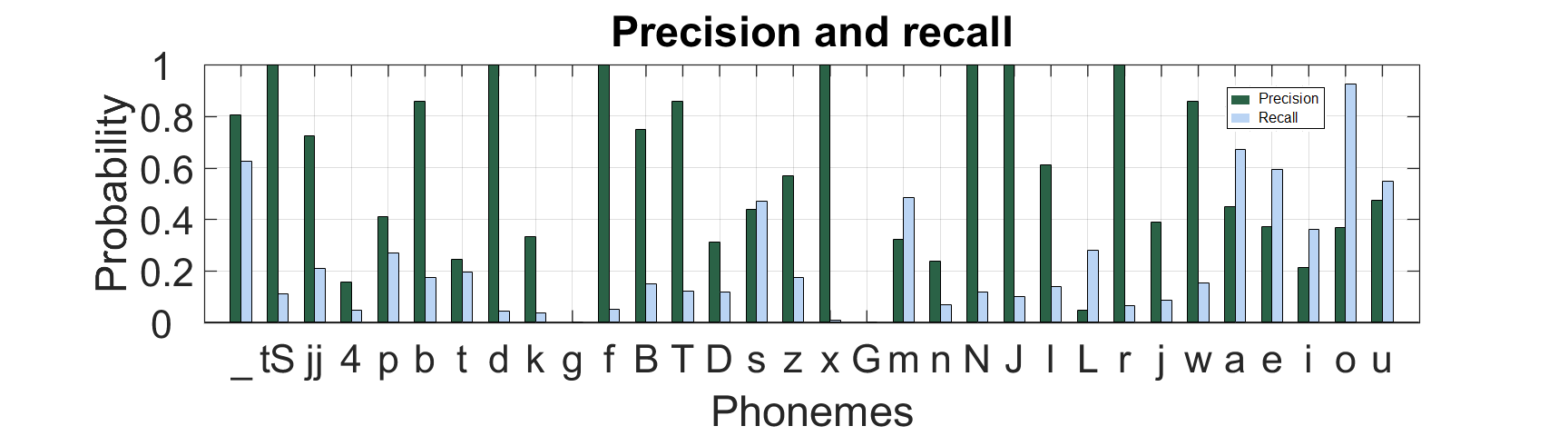}
        \caption{Top: Number of wrong detected phonemes. The red columns represent the false negatives phonemes and the green ones the false positives.; Bottom: Precision and Recall of each phoneme.}
        \label{fig:Misdetected}
\end{figure}

The word recognition rates reported by our system are rather low compared to those obtained by human observers. However, as stated earlier, our system is trying to recognize phonemes and convert them to words, so it is also interesting to analyze its performance in terms of phoneme recognition. The phoneme recognition rates obtained by the system are between 40\% and 60\%, as shown in Fig. \ref{fig:WaccAllUsersSystem} (\textit{Bottom}) and Fig. \ref{fig:SystemVShuman} (\textit{Bottom}). It is interesting to note that system performance was much more stable across participants than human performance. In addition, in terms of phoneme units, the global mean of the automatic system was 51.25\%, very close to the global mean of 52.20\% obtained by humans.

There are several factors that help understanding why the system achieves significantly higher rates in terms of phonemes than in terms of words: 1) Phoneme accuracy is computed at frame level because that is the output rate of the system. Thus, the temporal resolution used for phonemes is much higher than that of words and correctly recognizing a word implies the correct match of a rather long sequence of contiguous phonemes. Any phoneme mismatch, even if in a single frame, results in the whole word being wrong. 2) The automatic system finds it easier to recognize concrete phonemes (e.g: vowels) with high appearance rates in terms of frames (vowels are usually longer than consonants). This implies that a high phoneme recognition rate does not necessarily mean that the message is correctly decoded. To analyze this, system performance is displayed in Fig. \ref{fig:Misdetected}. Specifically, in Fig. \ref{fig:Misdetected} (\textit{Top}) we can observe the number of phonemes that were wrongly detected, distinguishing false negatives (in red color) and false positives (in green), while Fig. \ref{fig:Misdetected} (\textit{Bottom}) shows the corresponding values of precision and recall. Most of the consonants have very high precision, but many samples are not detected, deriving in a low recall. In contrast, vowels have an intermediate precision and recall because they are assigned more times than their actual occurrence. Close inspection of our data suggests that this effect is partially explained by the difficulty in correctly identifying the temporal limits of phonemes.

\section{Discussion and Conclusions}
\label{sec:Conclusions}

In this work we explore visual speech reading with the aim to estimate the recognition rates achievable by human observers and by an automatic system under optimal and directly comparable conditions. To this end, we recorded the VLRF database, appropriately designed to be visually informative of the spoken message. For this purpose we recruited 9 hearing-impaired and 15 normal-hearing subjects. Overall, the word recognition rate achieved by the 24 human observers ranged from 44\% (when the sentence was pronounced only once) to 73\% (when allowing up to 3 repetitions). These results are compatible to those from Duchnowski et al. \cite{duchnowski2000development}, who stated that even under the most favorable conditions (including repetitions) "speech-readers typically miss more than one third of the words spoken".

We also tested the performance of participants grouped by their hearing condition to compare their lip-reading abilities and verify if these are superior for hearing-impaired subjects, as suggested in some studies. Concretely, we found that hearing-impaired participants outperformed normal-hearing participants on the lip-reading task, but without statistical significance. The performance difference, which averaged 20\%, was not sufficient to conclude significance with the current number of subjects. Hence, future work will address the extension of the VLRF database so that it includes sufficient subjects to reach a clearer conclusion.

The participation of hearing-impaired people was very important given their daily experience in lip-reading. During the recording sessions they explained that lip-reading in our database was a challenge because they did not known the context of the sentence beforehand. For them, it is easier to lip-read when they know the context of the conversation. The conversation topic constrains the vocabulary that can appear in the talk. Furthermore, we mentioned before that lip-reading is related to the intelligence and the language knowledge. During the recording sessions we noticed that sentences directly related to daily life were easier to understand than sentences with words not used in colloquial language.

Another important aspect to consider is how easy or difficult is to lip-read different speakers. As explained in Section \ref{sec:VLRF}, participants were instructed to use their own criterion to facilitate lip-reading. It is difficult to objectively judge the effectiveness of the techniques that were used, but we observed some interesting tendencies during the recordings. Firstly, facial expressions help decoding the spoken message adding context to the sentence (e.g: sad expression if you are speaking about something unfortunate); hearing-impaired participants used this technique more often than normal-hearing subjects. Secondly, it is more useful to separate clearly between words than to exaggerate pronunciation. That is because the human system is searching words that fit the lip movements. We noticed that when pronunciation was exaggerated the separation between words was not clear or even lost considerably increasing the difficulty of lip-reading.

The above is important when interpreting the results of human observers, as they are conditioned both by the lip-reading abilities of the lip-reader and by the pronunciation abilities of the speaker. Recall that, in our experiments, each participant only lip-read his/her corresponding partner. It would be interesting to separate these factors, which could be done by randomizing the combinations of speakers and lip-readers on a per-sentence basis. In particular, the most interesting aspect would be to estimate the level of difficulty to lip-read each of the speakers, which could be done by having several subjects lip-reading the same speaker. There would be several advantages in doing so: 1) it would allow a more direct comparison to the performance of the system, as speaker performance will not be conditioned to a single human reader; 2) speakers that are too difficult could be excluded from the analysis, at least when seeking for the theoretical limit of lip-reading in optimal conditions; 3) it would help understand which are the best speaking techniques to use to facilitate lip-reading understanding.

As just explained, in our experiments, human observers reached word accuracy of 44\% in the first attempt while our visual-only automatic system achieved 20\% of word recognition rate. However, if we repeat the comparison in terms of phonemes, the automatic system achieves recognition rates quite similar to human observers, just above 50\%. These results are comparable with those reported by Lan et al. \cite{lan2012insights} who tested in the RM corpus, using 12 speakers and 6 expert lip-readers. Concretely, their human lip-readers reached 52.63\% viseme accuracy (in our case 52.20\% phoneme accuracy) and their system obtained 46\% viseme accuracy (our system 51.25\% phoneme accuracy). Therefore, in terms of viseme/phoneme accuracy, both Lan's and our system reach near-human performance. But this does not happen in terms of word accuracy: Lan et al. reported human word accuracy of 21\% (ours 44\%) and system word accuracy of 14\% (ours 20\%).

When trying to explain the above, we found that the low word recognition rates were related to: 1) the fact that it is quite easy to make mistakes at frame level and a mistake in a single frame results in the whole word being wrong; 2) the imbalance in the occurrence frequencies of phonemes. The latter is especially important because it highlights that the system, while achieving similar phoneme rates to those from humans, does not actually perform equally well. In other words, the phoneme sequences returned by humans always make some sense, which is not generally true for the system as it does not include higher-level constraints (e.g. at the word- or phrase-level). Hence, future directions should focus on introducing constraints related to bigger speech structures such as connected phonemes, syllables or words.

\section*{ACKNOWLEDGEMENTS}
This work is partly supported by the Spanish Ministry of Economy and Competitiveness under the Ramon y Cajal fellowships and the Maria de Maeztu Units of Excellence Programme (MDM-2015-0502), and the Kristina project funded by the European Union Horizon 2020 research and innovation programme under grant agreement No 645012.


\end{document}